# Exploring Anti-Aging Literature via *ConvexTopics* and Large Language Models.

Lana E. Yeganova, PhD, Won G. Kim, PhD, Shubo Tian, PhD, Natalie Xie, Donald C. Comeau, PhD, W. John Wilbur, MD PhD FACMI, Zhiyong Lu, PhD FACMI

Division of Intramural Research (DIR), NLM, NIH, Bethesda, MD USA 20894

**Abstract**

The rapid expansion of biomedical publications creates challenges for organizing knowledge and detecting emerging trends, underscoring the need for scalable and interpretable methods. Common clustering and topic modeling approaches such as K-means or LDA remain sensitive to initialization and prone to local optima, limiting reproducibility and evaluation. We propose a reformulation of a convex-optimization–based clustering algorithm that produces stable, fine-grained topics by selecting exemplars from the data and guaranteeing a global optimum. Applied to ~12,000 PubMed articles on aging and longevity, our method uncovers topics validated by medical experts. It yields interpretable topics spanning from molecular mechanisms to dietary supplements, physical activity, and gut microbiota. The method performs favorably, and most importantly, its reproducibility and interpretability distinguish it from common clustering approaches, including K-means, LDA, and BERTopic. This work provides a basis for developing scalable, web-accessible tools for knowledge discovery.

**Introduction**

The exponential growth of biomedical publications presents a critical challenge for researchers seeking to organize, interpret, and discover knowledge or emerging scientific trends (1, 2). Anti-aging research, in particular, has recently accelerated due to heightened public interest and a rapidly expanding commercial market for longevity interventions, many of which lack rigorous clinical evaluations. This gap between widespread adoption and scientific evidence underscores the need for systematic, data-driven analyses that yield reproducible results. In this work, we introduce *ConvexTopics*, a novel topic modeling algorithm, and demonstrate its utility by applying it to the rapidly expanding biomedical domain of anti-aging research.

Document clustering (3-5) and Topic modeling (6-8) are two common techniques for providing a broad overview of the information space and helping make sense of large text collections. Topic modeling identifies abstract themes or topics within a collection of documents by analyzing word co-occurrence, with each topic represented by a set of topic terms. Document clustering, on the other hand, groups documents into subsets based on content similarity. The two approaches are connected: topic terms can be derived from document clusters, and document clusters can be inferred from topic terms.

While there is a rich landscape of clustering and topic modeling approaches most share the same limitations: a) the user must specify the number of topics/clusters in advance, and b) methods generally converge to a local optimum, with the solution varying based on the choice of input parameters. Additionally, automatically evaluating their output and determining the optimal number of topics are longstanding challenges, with no effective automated solutions (9).

Lashkari and Golland (10) recently introduced a convex reformulation of likelihood functions for clustering tasks. Their approach restricts potential cluster centers to data exemplars, leading to a convex objective function and guaranteeing convergence to a global optimum. The number of resulting topics is determined automatically during computation rather than specified in advance. This algorithm has demonstrated strong performance on numerical datasets (sparse subset selection (11), or Multiview clustering (12)), however, its potential in natural language domains remains unexplored. In this paper, we advance the theoretical framework of the original paper (10) by reinterpreting it for natural language data and integrating the Dice coefficient as similarity measure. We refer to this adapted approach as *ConvexTopics* and emphasize that it is a convex method, which provides a globally optimal solution, and does not require the user to define the number of clusters.

We evaluate *ConvexTopics* on both general-purpose NLP benchmarks and biomedical text corpora. The annotated baselines include 20-Newsgroups and Reuters-21578, while the biomedical datasets consist of PubMed articles on *Anti-Aging*, *Diabetes Mellitus*, and *Age-Related Macular Degeneration*. We compare our method with state-of-the-art clustering and topic modeling approaches, including K-means (3, 4), LDA (6) and BERTopic (8). We introduce a new evaluation method, MaxMAP Topic Alignment, which is based on humanly assigned labels. Our experiments demonstrate that the method outperforms K-means and is competitive with LDA but much faster. Expert review confirmed that the topics produced by *ConvexTopics* align closely with accepted biomedical categories. This work introduces not only a robust clustering algorithm for large-scale text corpora but also lays a foundation for a web-based tool that can enable researchers to navigate large biomedical knowledgebases with transparency, reproducibility, and reduced dependence on heuristics in real time.

**Method**

*ConvexTopics* – **Convergence to a global optimum.** In this work, we adapt the theoretical framework presented in the paper (10), which approximates mixture fitting for clustering. The authors introduced an exemplar-based likelihood function which assumes that the centers of the individual distributions in a mixture can be well approximated by points within the sample itself. These sample points are called exemplars. This assumption simplifies the search space from an infinite, continuous one to a finite, discrete one. If the clusters are dense enough, this is a reasonable approximation, as a true cluster centroid will be very close to a data point. This restriction is a key part of what makes the resulting objective function convex.

Given a set of $n$ data points $X = \{x_1, \ldots, x_n\} \subset R^d$, $q_j$ is the probability that $x_j$ is a source point and $s_{ij}$ is the probability that $x_j$ could be the source of $x_i$. If we are dealing with Gaussians with known variance, we can, for any pair of points in our sample, compute the probability $s_{ij}$ that point $x_i$ came from source $x_j$.

Then $\sum_{j=1}^{n} s_{ij} q_j$ is the probability of seeing point $x_i$ and we can write the log likelihood of seeing all the data points as $\sum_{i=1}^{n} \log(\sum_{j=1}^{n} s_{ij} q_j)$. Intuitively, equation 1 describes the log-likelihood of seeing the data points, under a mixture-of-exponentials model. It is the log likelihood that we aim to maximize.

$$l(\{q_j\}_{j=1}^{n}; X) = \frac{1}{n} \sum_{i=1}^{n} \log \sum_{j=1}^{n} s_{ij} q_j = \frac{1}{n} \sum_{i=1}^{n} \log \sum_{j=1}^{n} q_j \, e^{-\beta \, d_\varphi(x_i, x_j)} + const \qquad (1)$$

In the formulation, the authors use the exponential family distribution on random variable X, which includes a broad family of distance measures, such as Euclidean distance or Kullback-Leibler (KL) divergence (13).

This is now a convex optimization problem and has been shown to be solvable by an efficient iterative algorithm with guaranteed convergence to the globally optimal solution (14).

$$z_i^{(t)} = \sum_{j=1}^{n} s_{ij} q_j^{(t)} \qquad \eta_j^{(t)} = \frac{1}{n} \sum_{i=1}^{n} \frac{s_{ij}}{z_i^{(t)}} \qquad q_j^{(t+1)} = \eta_j^{(t)} q_j^{(t)} \qquad (2)$$

Once the algorithm has converged, we compute soft term assignment probabilities to topics as follows:

$$r_{ij}^{(t)} = \frac{s_{ij} \, q_j^{(t)}}{z_i^{(t)}} \qquad (3)$$

In applying this method to the NLP field, the algorithm takes as an input a set of terms extracted from a set of documents. To achieve a rich text representation, we incorporate single terms minus stop words and noun phrases as the working vocabulary. We calculate a p-value for each candidate term using Hypergeometric distribution. It represents the probability that the term appears as often within the set of documents we are focused on, as opposed to the rest of PubMed, on a random basis. We adopt a useful statistical correction for multiple comparisons, the Benjamini–Hochberg correction (15) with False Discovery Rate set to 0,01, which allows one to achieve the desired confidence level for the family of terms. This allows us to prune the vocabulary we use in a practical way.

In our adaptation, $s_{ij}$ is the Dice coefficient (16) and is the probability that if one of the terms $i$ or $j$ is seen in a document, the other is also present. The Dice coefficient is widely used in NLP applications for set-based similarity, especially in word alignment, synonymy detection, and information retrieval (16). Our implementation defines $s_{ij}$ using term co-occurrence probabilities and has the benefit of being symmetric. We use a cut-off value for $s_{ij}$ and set it to zero if $s_{ij} < 0.05$, determined empirically to get the best results on the MeSH (Medical Subject Headings) data. This makes a sparse matrix, which greatly speeds up the computations and enables *ConvexTopics* to produce optimal number of clusters defined by $q_j > 0$. Terms with $r_{ij} > 0$ are assigned as relevant to topic $j$. Document score based on a topic is computed using the following formula:

$$score(d)_j = \sum_{i}^{n} r_{ij} * local\_weight_i(d) \qquad (4)$$

where local weight of term $i$ in document $d$ is determined from the document as defined in (17) and $r_{ij}$ is the weight of term $i$ in topic $j$, as defined in equation 3. We refer to this implementation of the algorithm as *ConvexTopics* emphasizing its convex nature and relation to topic modeling.

**Defining Number of Topics.** Topic models and clustering methods require the user to specify the number of topics beforehand. Researchers typically experiment with several choices of *K* and compare the resulting models. The selection is often made by manually reviewing the outputs and determining which one best supports the aims of the analysis. In a survey of 189 topic-modeling studies (18), authors point out that one of the main concerns across the studies is lack of methodological support for selecting the optimal number of topics *K*.

In contrast, the *ConvexTopics* algorithm determines the optimal number of topics as part of the computation itself, offering a significant methodological advantage. The formulation is computationally efficient, and while fixing *K* would normally introduce a combinatorial problem – choosing *K* cluster centers from among *n* data points – the convex framework resolves this by allowing every data point to act as a potential cluster-center candidate. This eliminates the need to search over discrete subsets of exemplars and enables globally optimal clustering solutions.

**MaxMAP Topic Alignment Evaluation.** To evaluate the quality of the topics produced by *ConvexTopics*, we leverage MeSH term assignments as a source of expert-curated labels. We designed an evaluation framework that measures how well computed topics align with biomedical concepts defined by MeSH. MeSH is the controlled vocabulary used for indexing biomedical literature [15], and the terms assigned to a PubMed article reflect its primary topics. As nearly all PubMed articles receive 10–15 MeSH terms [20], the biomedical domain offers a uniquely rich resource for large-scale evaluation. MeSH terms have been widely used for evaluating information retrieval, topic modeling and clustering systems (19-22), and they provide a natural foundation for assessing topic assignments.

Topic-modeling and clustering methods rely either on automated metrics that correlate weakly with human preferences or on expert labels that do not scale [6, 19]. Traditional evaluation focuses on topic coherence—typically whether top-ranked words form a recognizable concept—which is often approximated using automated measures such as normalized pointwise mutual information (NPMI). However, coherence metrics have repeatedly been shown to align poorly with human judgments [6] and measure only how closely top terms cluster together, not whether a topic meaningfully organizes documents.

To address these limitations, we introduce MaxMAP Topic Alignment, a MeSH-driven evaluation method that directly measures how well each topic captures an expert-defined biomedical concept. For each dataset, we collect all MeSH terms assigned to its documents and filter them using the same procedure as filtering the vocabulary for topic terms for *ConvexTopics* (Hypergeometric p-values and Benjamini–Hochberg correction). For each qualified MeSH term, we compute its Average Precision (AP) across topics and assign the topic with the highest AP score to the MeSH term. In each topic the document scores are used to rank the documents for computing the Average Precision. We repeat this process for all qualified MeSH terms, and then select the *N* MeSH terms with highest AP values and compute their average as Mean Average Precision (MAP). We use this MaxMAP Topic Alignment measure to compare *ConvexTopics* with other approaches. Unlike coherence scores, our metric directly measures alignment with expert curation.

Intuitively, the MaxMAP evaluation measures how effectively a topic ranks documents associated with a given MeSH label. To illustrate this alignment, we provide several examples from the Anti-Aging dataset where *ConvexTopics* identifies topics that correspond closely to specific MeSH concepts. For instance, the MeSH term *isoptera / metabolism* corresponds to a topic whose exemplar term is "social insects," reflecting biological studies on termite metabolism. Similarly, *fallopia japonica / chemistry* aligns with a topic exemplified by "cuspidatum," the plant's taxonomic name. Other strong examples include *cordyceps / genetics*, which maps to a topic defined by the exemplar

"cordyceps"; *picrorhiza / chemistry*, which corresponds to a topic centered on "rhizomes"; and *keratosis, actinic*, which maps directly to a topic with the exemplar "keratosis."

**Experiments and Results**

**Datasets**. We analyze the performance of the algorithm on two groups of datasets. The first group includes well established benchmarks widely accepted by the community – Reuters RCV1 dataset (23) and the 20 Newsgroups dataset (http://people.csail.mit.edu/jrennie/20Newsgroups). The Reuters-21578 corpus is a collection of newswire articles originally distributed by Reuters for research in text classification. It contains thousands of manually labeled documents spanning diverse economic and political topics, making it a standard dataset for evaluating classification and clustering methods. 20-Newsgroups corpus is a collection of approximately 20,000 Usenet posts evenly distributed across 20 different discussion groups. The categories cover a broad range of topics, including politics, religion, science, sports, and technology, with some closely related groups and others more distinct.

The second group are biomedical datasets collected from PubMed. For our experiments, we have selected 3 datasets that provide a variety in terms of size and complexity: the *Anti-Aging* dataset, the *Diabetes Mellitus* dataset, and the *Age-related Macular Degeneration* dataset. The Anti-Aging dataset contains 12,167 PubMed documents and is created using PubMed query [*anti-aging or antiaging or "anti-aging"*]. The *Diabetes Mellitus* dataset (192,333 PubMed documents) is created using PubMed query "Diabetes Mellitus type 2 [MeSH]". The *Age-related Macular Degeneration dataset (*23,030 PubMed documents*) is created using PubMed query "Age-related Macular Degeneration"* and is limited to documents from the last 10 years.

**Models.** We compare *ConvexTopics* with K-means (with *tf x idf vectors*), a popular baseline clustering method, LDA (bag-of-words local count, Gensim implementation with stopping criteria), a popular baseline topic modeling method. Additionally, we compare our algorithm with BERTopic, a neural topic model, and use the following default settings: *sentence-transformers* as an embedding model, *UMAP* for dimension reduction, and *HDBSCAN* for clustering, with concatenation of title and abstract as input. Recent studies (24) indicate that a classical LDA model performs at least as well, if not better, than its modern equivalents, and thus presents a strong baseline.

**Evaluation**. Each topic produced by *ConvexTopics* is characterized and is interpretable through a representative set of single terms or noun phrases, defining the topic. These terms are used to score documents in regard to the topic, as outlined in equation 4. Hence, each topic provides a different ordering of documents, defined by topic terms in a topic.

*ConvexTopics* generates 2,194 topics for Anti-Aging set, 5,345 topics for Diabetes Mellitus set, and 2,670 topics for Age-related Macular Degeneration set. They all took less than a minute to execute, using 20 threads in our multithread implementation. *ConvexTopics* often generated more than 1,000 topics, in which case we use the best 1,000 for evaluation. Additionally, in our MaxMAP evaluation, we use the 1,000 MeSH terms with the highest Average Precision to compute the MAP. Accordingly, we use 1,000 as input for LDA and K-means to enable direct comparison with *ConvexTopics*. BERTopic does not require specifying the number of topics and instead uses a minimum cluster size, which we set to 10.

*ConvexTopics* generates 1,248 topics for the Reuters dataset and 548 for 20-Newsgroups. We additionally compute topics/clusters for LDA and K-means using the number of classes specified in the gold-standard datasets—120 for Reuters and 20 for 20-Newsgroups—to provide another comparison point.

| MaxMAP Evaluation | Reuters-RCV1 | 20-Newsgroups |
| --- | --- | --- |
| *ConvexTopics* (# of topics) | 0.2506 (1,000) | 0.2759 (548) |
| LDA (# of topics) | 0.2291 (1,000) // 0.0826 (120) | 0.2673 (548) // 0.2945 (20) |
| K-means (# of clusters) | 0.1637 (1,000) // 0.0904 (120) | 0.1403 (548) // 0.2108 (20) |
| BERTopic (# of clusters) | 0.164009 (330) | 0.161891 (221) |

Table 1. MaxMAP Topic Alignment Evaluation on Reuters-RCV1 and the 20 Newsgroups datasets.

Table 1 shows that *ConvexTopics* achieves the highest MaxMAP scores across both benchmark datasets, outperforming LDA, K-means, and BERTopic when setting *N* to 1,000 for Reuters and 548 for 20-Newsgroups, as guided by *ConvexTopics*. For Reuters, *ConvexTopics* reaches a MaxMAP of 0.2506, outperforming LDA (0.2291) and substantially surpassing K-means (0.1637) and BERTopic. Similarly, for 20-Newsgroups, *ConvexTopics* reaches a MaxMAP of 0.2759, outperforming LDA, K-means, and BERTopic. The results remain consistent when using the number of topics specified by the number of labels in Reuters-RCV1 (120 topics) and 20-Newsgroups (20 clusters), except for LDA on 20-Newgroups. Note that this comparison is not fully fair to K-means, since it produces a partition of documents into hard clusters, whereas LDA and *ConvexTopics* produce topic-specific term weights that naturally induce a ranking of all documents for each topic. Soft K-means is an option to be considered. We have done some preliminary computations, showing that one can make an improvement over the numbers in our table. However, the computations are lengthy, thus far making it impractical for our purposes.

| MaxMAP Evaluation | Anti-Aging | Diabetes Mellitus | Age-related Macular Degeneration |
| --- | --- | --- | --- |
| *ConvexTopics* | 0.4850 | 0.4810 | 0.4215 |
| LDA | 0.3744 | 0.2660 | 0.3539 |
| K-means | 0.3470 | 0.1305 | 0.2184 |
| BERTopic | 0.3372 (188) | 0.1233 (1812) | 0.1998 (310) |

Table 2. MaxMAP Topic Alignment Evaluation on Anti-Aging, Diabetes Mellitus, and Age-related Macular Degeneration datasets.

Similarly, Table 2 shows that *ConvexTopics* achieves the highest MaxMAP scores across all three biomedical datasets—Anti-Aging, Diabetes Mellitus, and Age-related Macular Degeneration. *ConvexTopics* obtains MaxMAP values of 0.4850, 0.4810, and 0.4215, respectively, consistently outperforming LDA, K-means, and BERTopic.

**Application to *Anti-Aging* literature in PubMed**

The pursuit of longevity, anti-aging, and healthy living has generated a vast array of online resources promoting diverse supplements, interventions, and lifestyle practices. While many people actively incorporate these practices into their daily routines, some of the publicly available information is not grounded in rigorous scientific evidence. This gap highlights the need for systematic organization and analysis of the biomedical literature on aging to identify which claims are supported by research. *For example,* how do physical exercise, body composition, and gut microbiome influence aging and longevity? Are $NAD^+$ supplements safe and effective for increasing energy levels? Are there clinical trials validating Ayurvedic treatments? What evidence supports the use of Royal Jelly? What are the benefits and risks of dermatologic procedures such as laser therapy or fillers? Addressing such questions requires accessible, well-structured summaries of the underlying science.

$NAD^+$ supplements, for instance, are widely available on the market, however, there is insufficient evidence regarding their safety. The studies indicate that there are many unknowns regarding pharmacokinetics and pharmacodynamics, particularly bioavailability, metabolism, and tissue specificity of $NAD^+$ boosters (25). Studies also indicate the lack of long-term safety studies, and need for more clinical trials to determine the proper dose of $NAD^+$ boosters. They have been reported in PubMed studies to potentially increase kidney inflammation in older population (26, 27).

**Corpus.** Out Anti-Aging corpus contains ~12,000 PubMed articles satisfying the query [*anti-aging or antiaging or "anti-aging"*]. This is the same corpus as the one used for evaluation. We took all the terms that satisfy our criteria, and that was 31,047 terms. For this collection, *ConvexTopics* generates 2,194 focused topics. Each topic includes:
1. Topic Terms — terms that define the topic computed by *ConvexTopics.*
2. Top scoring PMIDs — the most relevant PubMed articles on this topic (top 30 are presented on the website). The documents are selected based on their relevance to the topic terms using equation 4. Clicking on a PMID leads to the article view in PubMed, where one can read, download, or find links to full-text articles if available.
3. Topic Summary and Title — a short paragraph and title describing and summarizing the topic. Summaries and titles are generated using GPT-4o applied to the top ten scoring documents associated with the topic. Manual evaluation of a select set of topics confirmed that these summaries are mostly well grounded to references, offering informative descriptions of each topic.

A large number of focused and reproducible topics is a strength of *ConvexTopics*, but such granularity can be difficult for humans to browse and interpret. To provide a more accessible overview, we use LLMs to group the fine-grained topics into higher-level thematic categories, offering a landscape view of the research while preserving the algorithmically derived structure.

**Navigating the Web Site**: We have created a web interface serving as a gateway to exploring PubMed articles related to anti-aging https://www.ncbi.nlm.nih.gov/CBBresearch/Wilbur/IRET/topics/antiaging/. The main page displays these higher-level topics, each represented by a short descriptive title. Using this website, one is guided from high-level major categories to lower level *ConvexTopics* associated with that research area, enabling one to move from broad themes to fine-grained topics. Figure 1 shows high-level topics presented on the main page.

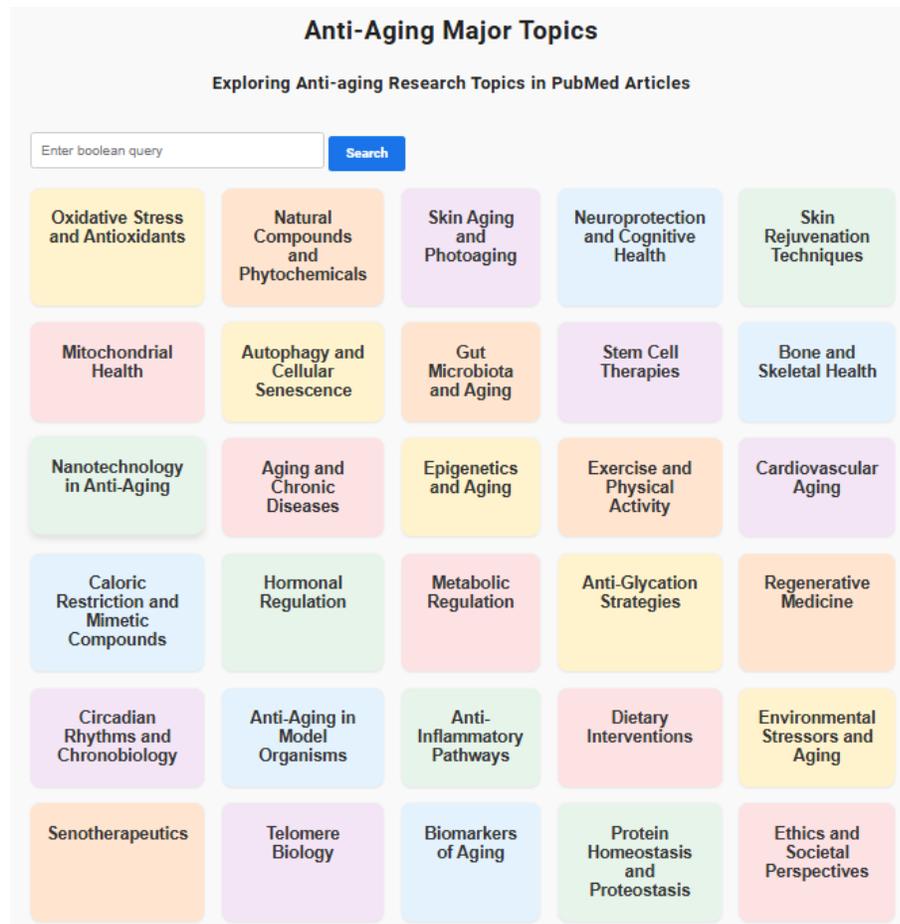

Figure 1: High-level topics presented on the main page of Anti-Aging Website.

The website provides an opportunity to search the topics using any key words. The search can be conducted on the main page and will retrieve all focused topics that are relevant to the search term. Additionally, search functionality is implemented within each high-level topic and allows one to search and retrieve *ConvexTopics* topics that are in that subspace. Here we provide an expert review of two of the topics. One topic is Sarcopenia and appears under the *Exercise and Physical Activity* theme. The second topic explores Gut Microbiota and its relation to aging.

**Sarcopenia**. The *ConvexTopics* algorithm produced a coherent cluster centered on sarcopenia, the age-associated decline in skeletal muscle mass, strength, and functional performance. Expert review highlighted that contemporary sarcopenia research integrates mechanistic biology, diagnostic frameworks, cardiogeriatric implications, and emerging therapeutic strategies, reflecting the multidimensional nature of this geroscience condition. Three key publications illustrate the scientific depth and heterogeneity captured within this cluster.

A recent mechanistic study by Li et al. (28) compared aging trajectories in limb muscles versus the diaphragm, demonstrating that limb skeletal muscles undergo classic sarcopenic changes—atrophy, impaired autophagy, loss of strength—whereas the diaphragm remains structurally and functionally resilient with age. Transcriptomic profiling identified 61 genes altered in aging limb muscle but not in the diaphragm; among these, Smox (spermine oxidase) emerged as a modulator of mitochondrial integrity and muscle mass. Experimental manipulation of Smox improved mitochondrial function and muscle performance, suggesting that diaphragm-like "anti-aging programs" may reveal novel therapeutic targets for limb muscle sarcopenia.

At the clinical interface, Anagnostou et al. (29) synthesized evidence linking sarcopenia with cardiovascular aging, emphasizing a strong bidirectional interaction between muscle decline and cardiovascular disease (CVD). Sarcopenia accelerates adverse outcomes across multiple conditions, including heart failure, aortic stenosis, peripheral artery disease, and atrial fibrillation, while CVD-related factors such as neurohormonal activation, oxidative stress, and inactivity further exacerbate muscle loss. Diagnostic criteria emphasize loss of muscle strength, especially handgrip, as the first indicator of sarcopenia, with confirmation by reduced muscle mass and further grading by walking speed. Measures such as grip strength and walking speed consistently outperform imaging-based assessments for predicting health risks, highlighting their importance in both geriatrics and cardiology.

A broader overview by Liu et al. (30) highlights advances in molecular mechanisms and diagnostics, presenting sarcopenia as a disorder arising from convergent declines in anabolic signaling (IGF-1/Akt/mTOR), increased catabolic drive (FOXO-UPP, myostatin/SMAD), chronic inflammation (NF-κB), and impaired mitochondrial biogenesis and mitophagy. Current best-practice management remains multimodal, with progressive resistance training (2–3×/week) as the therapeutic foundation, supported by optimized protein intake (1.2–1.6 g/kg/day), vitamin D repletion, and targeted management of endocrine, metabolic, cardiovascular, and inflammatory comorbidities. Emerging gerotherapeutic approaches—including NAD+ precursors, AMPK/mTOR modulation, myostatin inhibitors, senolytics, and autophagy enhancers—show mechanistic promise but require rigorous validation, especially in older adults with CVD.

Together, these studies demonstrate that the *ConvexTopics* sarcopenia cluster corresponds to a clinically important and therapeutically evolving domain of aging biology. The alignment between algorithmic clusters and expert-validated themes—muscle-specific aging trajectories, cardiogeriatric interactions, and convergent molecular drivers—illustrates the utility of convex clustering for extracting domain-relevant insights from large biomedical corpora.

**Gut Microbiota and Aging.** We have been interested in gut microbiome and its possible significance in prolonging human life and health span. Clicking on *Gut Microbiota and Aging* major topic brought us to a page with 32 focused topics computed by our algorithm. Here we found the topic *Intestinal Barrier Integrity: Key to Aging and Cognitive Health* of interest and on selecting it we were taken to a page with a summary of the ten top documents. Authors in (31) discuss *inflammaging* , a chronic low grade inflammation not due to infection per se. It is increased with poor health and increasing age and is generally believed to be an important contributing factor in aging. It is present in the gut as an organ and becomes more marked as the wall of the gut becomes more permeable with age and toxins and even bacteria more easily penetrate the wall and trigger an immune reaction. Evidence is presented that "gut microbiome composition, gut immune function, and gut barrier integrity are involved in the formation of inflammaging in the aging gut."  This paper leaves open what is the major driver of inflammaging in the gut.

One of the ways the gut microbiome has been manipulated is through fecal microbiota transplant (FMT). In this way the organisms present in the gut of one animal can be transplanted into another animal. FMT has been successfully used in humans to cure cases of Clostridium difficile infection. Under the topic *"Metabolomics: Microbiome and Metabolic Therapies for Anti-Aging"* we found two related studies (32) and (33) that provide an interesting evidence that the organisms growing in the gut of an animal can cause benefit or harm and strongly suggests that the organisms are an important factor in the health of an animal. However, the results of FMT are generally short lived and it seems likely something else is needed to prolong a healthy life.

A longer term way to manipulate the gut microbiota is through the regular administration of probiotics. Five of the topics involve probiotics in one way or another and we looked at all these topics and almost all the research is on model organisms. There are some results showing the benefit of probiotics for human disease states but nothing that shows prolongation of a healthy life span.

Almost all the evidence we have mentioned so far is in model organisms not humans. Under the topic *"Advances in Extending the Human Lifespan: Genes, Therapies, and Microbiota"* we found a paper discussing calorie restriction and its relation to lifespan extension (34). This paper points out that calorie restriction has been shown to prolong healthy human life, but humans find it quite difficult to stay on a regimen involving calorie restriction. As a solution it has been proposed that certain drugs produce effects in the body mimicking calorie restriction (CRM, calorie restriction mimetic). Such drugs may provide a similar benefit to calorie restriction. Both calorie restriction and the effects of CRMs change the gut microbiome, and it is theorized that the changes in the gut microbiome may be the source of at least a portion of their benefit. So far this remains an interesting hypothesis.

It is a little disappointing that there is yet no clear path to prolonging a healthy human life span by manipulating the gut microbiome. However, in studying the paper on CRMs we were pointed to two papers proving that knowledge of the bacteria growing in the human gut allow the prediction of an early death (35) or the prolongation of life (36). These results suggest that some benefits should be possible for humans through manipulation of the gut microbiome. For now, perhaps the best thing we as humans can do is to follow conventional wisdom and eat a naturally high fiber diet which is believed to provide a longevity benefit though we do not know how much of this benefit may be through promotion of a healthier gut microbiome.

**Conclusions and Future Work**

This study demonstrates the effectiveness of the *ConvexTopics* algorithm in overcoming key limitations of traditional clustering methods, including susceptibility to local optima and the need to predefine the number of topics. By formulating clustering as a convex optimization problem, the algorithm converges to a globally optimal solution, automatically determines the number of topics, and executes rapidly enough to support real-time applications. We further introduce a scalable evaluation framework based on MeSH terms.

We apply *ConvexTopics* to the rapidly expanding field of anti-aging research, uncovering hundreds of fine-grained topics. Each topic includes representative terms and articles, and we integrate GPT-4o to generate topic titles and concise summaries. We further aggregate these fine-grained topics into broader, high-level themes and link each focused topic to its corresponding high-level category.

As LLMs become increasingly integrated into biomedical text mining, their ability to generate user-friendly summaries is highly valuable; however, they remain prone to hallucination and inconsistent outputs. *ConvexTopics* provides a stable, reproducible backbone that constrains the use of LLMs, ensuring that summaries are grounded in algorithmically derived topics. We believe this controlled integration provides the benefit of human-readable summaries while minimizing hallucinations by keeping the LLM tightly anchored to the topic terms and relevant documents computed by *ConvexTopics*.

One of the limitations of the method is that it frequently generates a large number of narrowly focused clusters and we resort to ChatGPT to produce higher level topics. Also, *ConvexTopics* can produce a few large clusters based on exemplar terms that are not very meaningful. Preliminary review of the summaries generated by ChatGPT suggests that while the entities and relationships in summaries are well preserved, the sentiments and details may be misleading. This needs to be further quantified, though we find the results useful.

This work lays a foundation for building a web-based tool that allows researchers to apply this technique to any collection of biomedical articles in PubMed in real time. Such a tool would enable users to navigate biomedical knowledge with greater transparency, reproducibility, and reduced dependence on heuristic initialization.

**Acknowledgement.** This research was supported by the Division of Intramural Research (DIR) of the National Institutes of Health (NIH). The contributions of the NIH author(s) are considered Works of the United States Government. The findings and conclusions presented in this paper are those of the author(s) and do not necessarily reflect the views of the NIH or the U.S. Department of Health and Human Services.

The authors thank Elena Detwiler for valuable discussions regarding the website and for her input on the topic of *Exercise and Physical Activity*, particularly her insights on *sarcopenia*.